# Comparing Neural- and N-Gram-Based Language Models for Word Segmentation

**Yerai Doval**
*Grupo COLE, Departamento de Informática E.S. de Enxeñaría Informática, Universidade de Vigo Campus As Lagoas, Ourense 32004, Spain. E-mail: yerai.doval@uvigo.es*

**Carlos Gómez-Rodríguez**
*Grupo LYS, Departamento de Computación Facultade de Informática, Universidade da Coruña, Campus de Elviña, A Coruña 15071, Spain. E-mail: cgomezr@udc.es*

**Word segmentation is the task of inserting or deleting word boundary characters in order to separate character sequences that correspond to words in some language. In this article we propose an approach based on a beam search algorithm and a language model working at the byte/character level, the latter component implemented either as an n-gram model or a recurrent neural network. The resulting system analyzes the text input with no word boundaries one token at a time, which can be a character or a byte, and uses the information gathered by the language model to determine if a boundary must be placed in the current position or not. Our aim is to use this system in a preprocessing step for a microtext normalization system. This means that it needs to effectively cope with the data sparsity present on this kind of texts. We also strove to surpass the performance of two readily available word segmentation systems: The well-known and accessible Word Breaker by Microsoft, and the Python module WordSegment by Grant Jenks. The results show that we have met our objectives, and we hope to continue to improve both the precision and the efficiency of our system in the future.**

## Introduction

The concept of *word*, which we define here as a sequence of characters delimited by special word boundary characters, is truly important in natural language processing. There are several tasks and systems in this field that rely on word level information to achieve their goals. For instance, tokenization, which is a common preprocessing stage in a wide range of systems, usually requires that words are correctly delimited by blank characters in order to identify the corresponding tokens correctly. In this context,[1] a *token* is usually a word or a group of words that constitutes the most basic element to process in a particular task, such as in entity recognition (Tjong Kim Sang & De Meulder, 2003), POS tagging (Ratnaparkhi, 1996), or sentiment analysis (Vilares, Alonso, & Gómez-Rodríguez, 2017).

Although an English speaker can easily discern the words in "thepricewasfair," a machine can only see a sequence of characters—or more precisely, bytes—and may at first distinguish just one long word. By explicitly executing a *word segmentation* procedure, the machine inserts word boundary characters between sequences of characters that would end up corresponding to words in some particular language. The resulting text, in our example "the price was fair," can now be further processed in a word-by-word basis, as these elements are now clearly isolated from each other. It is worth noting that this scheme also works for incorrectly segmented texts such as "th e pricew asf air" by first removing all the word boundaries. In Table 1 we show some example instances of the problem we are trying to solve.

Our proposed approach for word segmentation uses a beam search algorithm aided by a byte or character level language model, which is implemented using a neural network or an n-gram model, respectively. The beam search algorithm processes the input one token at a time (a byte or character) generating, for each step, a set of partial segmentation candidates by checking the likelihood of the current candidates and the probability of the next token corresponding to a word boundary. This information is retrieved from the language model. Then, at the end of each step the *n* best candidates are chosen as input for the next step.

---

[1] Not in our case, where a token is either a byte or a character.

TABLE 1. Example instances of the problem we are trying to solve as input/output pairs.

| Input | Output |
| --- | --- |
| mostvaluablepuppets | most valuable puppets |
| webdesign | web design |
| RantsAndRaves | Rants And Raves |
| thankU | thank U |
| work allday | work all day |
| o m g u serious?? | omg u serious?? |
| Safe way is very rocknroll tonight | Safeway is very rock n roll tonight |

The main objective of our work is to devise a word segmentation system that can be used as a preprocessing step for a microtext normalization system (Doval, Vilares, & Gómez-Rodríguez, 2015). In this context, the biggest challenge comes from the type of texts we are dealing with, which are affected by *texting* phenomena such as character repetition (for instance, "hiiii" for "hi") or phonetic–based character substitution (for instance, "dawg" for "dog"), to name just a few (Thurlow & Brown, 2003). These phenomena introduce a great amount of data sparsity in the problem at hand, as "hiiii" is not *exactly* the same as "hii" or "hi" but must be treated equally by the segmenter as one word. This kind of texts are abundant in popular social media networks and microblogging platforms such as Twitter or Facebook, where large amounts of nonstandard textual data are created every second (at the time of this writing, http://www.internetlivestats.com/one-second/ reports that 7,602 new tweets are published every second). As it would be highly improbable to observe every standard word affected by every single type of texting phenomena, we abandon the word-level processing of these texts and opt instead for a character or byte level approach in order to tackle the resulting data sparsity problem. Furthermore, neural language models may also help in this regard thanks to the internal continuous representations they construct of their inputs. These representations, usually real-valued vectors, encode similarities between different input instances (Kim, Jernite, Sontag, & Rush, 2015), which can be later exploited in the task at hand. This contrasts with the discrete treatment of a word-based n-gram model, where every possible input element is unrelated to each other. Fortunately, a character-based n-gram model may be able to overcome this issue due to its finer-grained processing, the reason why we consider this type of model in this work.

In our experiments we compared the described approach with the Microsoft Word Breaker (Wang, Thrasher, & Hsu, 2011) and the WordSegment Python module by Grant Jenks.[2] The languages considered for our tests were English, Spanish, German, Turkish, and Finnish, and we also included a test set comprised of English tweets. The latter three languages are known for their complex morphology, with Turkish and Finnish being agglutinative languages and thus constituting a greater challenge for a segmentation system (Manning, Raghavan, & Schütze, 2008, Chapter 2).

Overall, our approach was able to outperform both the Word Breaker and WordSegment for all of the languages considered, with the sole exception of a tie with WordSegment in one of the Spanish data sets. But most notably for us, our systems obtained notable improvements in such an interesting case as the Twitter data set. Looking at the performance obtained by the different types of language models used, we surprisingly see strong numbers for the simpler and faster n-gram model, which was in several cases on a par with the more sophisticated neural model.

**Related Work**

Word segmentation is an important preprocessing step in several natural language-processing systems, such as machine translation (Koehn & Knight, 2003), information retrieval (Alfonseca, Bilac, & Pharies, 2008), or speech recognition (Adda-Decker, Adda, & Lamel, 2000). On the other hand, most Asian languages, although retaining the concept of *word*, do not use word boundary characters in their writing systems to separate these elements. As a result, the application of word segmentation for these languages has drawn a lot of attention from the research community, with abundant work in recent years (Chen, Qiu, Zhu, Liu, & Huang, 2015; Pei, Ge, & Chang, 2014; Xu & Sun, 2016; Zheng, Chen, & Xu, 2013).

Beyond the Asian context, we can also find European languages with highly complex morphology such as German, Turkish, or Finnish, which can also benefit from a conceptually different word segmentation procedure (Alfonseca et al., 2008; Koehn & Knight, 2003). In these cases, and mainly for agglutinative or compounding languages (Krott, Schreuder, Harald Baayen, & Dressler, 2007), new words are usually created just by joining together previously known words. A system with a vocabulary lacking these new words may still be able to process them if some sort of word segmentation system is in place. However, it is worth noting that this is a slightly different kind of word segmentation, as it is concerned with extracting the base words that form a compound word. In contrast, our approach focuses on separating all words, compound or not, from each other.

Moving on to the web domain, there are special types of tokens that can also be targeted by a segmentation system. The first ones to appear, and an essential concept for the web itself, are URLs (Chi, Ding, & Lim, 1999; Wang et al., 2011). These elements do not admit literal whitespaces in their formation, but most of the time they do contain multiple words in them. Words may be separated by a special encoding of the whitespace character like percent-encoding or a different encoding that uses URL-safe characters. Most other times, words are just joined together with no boundary characters, and thus the requirement for a segmentation process arises.

---

[2] http://www.grantjenks.com/docs/wordsegment/

Then, with the advent of Web 2.0, the use of special tokens called *hashtags* in social media became very common (Maynard & Greenwood, 2014; Srinivasan, Bhattacharya, & Chakraborty, 2012). Similar to URLs, hashtags may also be formed by multiple words. Unlike those, these elements do not use any word boundary character(s) between words, thus the use of a segmentation system seems more advantageous in this case.

The segmentation procedure that most of the previous work follows can be summarized in two steps. First, they scan the input to obtain a list of possible segmentation candidates. This step can be iterative, obtaining lists of candidates for substrings of the input until it is wholly consumed. Sets of predefined rules (Koehn & Knight, 2003) or other resources such as dictionaries and word or morpheme lexicons (Kacmarcik, Brockett, & Suzuki, 2000) may be used for candidate generation. Then, for the second step they select the best or $n$ best segmentation candidates as their final solution. In this case, they resort to some scoring function, such as the likelihood given by the syntactic analysis of the candidate segmentations (Wu & Jiang, 1998) or the most probable sequence of words given a language model (Wang et al., 2011).

Some other techniques, usually employed in the Chinese language, consider the word segmentation task as a tagging task (Xue, 2003). Under this approach, the objective of the segmentation system is to assign a tag to each character in the input text, rendering the word segmentation task as a sequence labeling task. The tags mark the position of a particular character in a candidate segmented word, and usually come from the following set: Beginning of word, middle of word, end of word, or unique character word.

Recently, neural network-based approaches have joined traditional statistical ones based on Maximum Entropy (Low, Ng, & Guo, 2005) and Conditional Random Fields (Peng, Feng, & McCallum, 2004). These models may be used inside the traditional sequence tagging framework (Chen, Qiu, Zhu, & Huang, 2015; Pei et al., 2014; Zheng et al., 2013) but, more interestingly, they also enable new approaches for word segmentation. Cai and Zhao (2016) obtain segmented word embeddings from the corresponding candidate character sequences and then feed them to a neural network for scoring. Zhang, Zhang, and Fu (2016) consider a transition-based framework where they process the input at the character level and use neural networks to decide on the next action given the current state of the system: Append the character to a previous segmented word or insert a word boundary. Both of these approaches use recurrent neural networks for the segmentation candidate generation and beam search algorithms to find the best segmentation obtained.

Outside the Chinese context, one of the most popular state-of-the-art systems for word segmentation in multiple languages is the Microsoft Word Breaker from the Project Oxford (Wang et al., 2011). Its original article defines the word segmentation problem as a Bayesian Minimum Risk Framework. Using a uniform risk function and the Maximum a posteriori decision rule, they define the a priori distribution, or segmentation prior, as a Markov n-gram. For the a posteriori distribution, or transformation model, they consider a binomial distribution and a word length-adjusted model. Finally, they solve the optimization problem posed by the decision rule using a word synchronous beam search algorithm.

The language model they use for the a priori distribution is presented in Wang, Thrasher, Viegas, Li, and Hsu (2010). This is a word-based smoothed backoff n-gram model constructed using the CALM algoritm (Wang & Li, 2009) with the web crawling data of the Bing search engine.[3] Some particular features of this model are that all the words are first lowercased and their non-ASCII alphanumeric characters transformed or removed to fit in this set, and also that it is being continuously updated with new data from the web. However, the aggressive preprocessing performed by this system may result in limitations in two particular domains: Microtexts and non-English languages. For the first case, data sparsity may pose a problem for a word-based n-gram language model. This type of model would have to see every possible variation of a standard word in order to process it appropriately. As an example, an appearance of the unknown word "hii" would mean using the ¡UNK¿ token instead of the information stored for the equivalent standard known word "hi," which constitutes some loss of information. Then, if the also unknown word "theeere" occurs, it would mean that the system has failed to use any relevant information to process the input. Hence, the input "hiitheeere" could be incorrectly segmented into "hii thee ere," a more likely path for the model given the known token "thee."

On the other hand, working only with lowercased ASCII alphanumeric characters leaves non-Latin alphabets out of the question—although Latin transcriptions could be used—and limits the overall capacity of the system due to the loss of information from the removed or replaced characters. For instance, consider "momsday" in the context of text normalization. The n-gram model would give higher likelihood to "mom s day" as it would have seen the token "mom" very frequently, both when appearing on its own and when swapping the "'" by a word boundary character in "mom's," obtaining "mom s day." However, we prefer in this case "moms day" as the most likely answer, not only because it can be the correct answer but also because we can later correct the first word to include the apostrophe if needed and/or appropriate using a text normalization system.

The WordSegment Python module is an implementation of the ideas covered in Norvig (2009). It is based on 1-gram and 2-gram language models working at the word level that are paired with a Viterbi algorithm for decoding.

---

[3] https://www.bing.com/

The system first obtains segmentation candidates that are scored using the n-gram models, and then the best sequence of segmented words is selected using the Viterbi algorithm. A clear advantage of this system for our work is that we can easily train its n-gram models from scratch in order to adapt it for our text domains/languages. This provides us with a better comparative framework than the Word Breaker.

Our current take on the word segmentation task extends the work in Doval, Gómez-Rodríguez, and Vilares (2016) with a new beam search algorithm and newer implementations for the language model component. We also broaden the scope of our work by targeting not only Spanish but also English, German, Turkish, and Finnish. We have chosen the last three languages based on the need to test our approach with morphologically complex languages, with the agglutinative languages Turkish and Finnish being the most notable cases.

## System Description

Before going into details, it is important to note that we will view the input text as a sequence of bytes when using the neural model and as a sequence of characters when using the n-gram model. We will refer to either a byte or a character as a *token*.

Our proposed approach is formed by two components: The beam search algorithm and the language model.

The search algorithm acquires the input text and first removes all word boundary tokens. Then it analyzes the resulting text one token at a time, deciding whether a word boundary token would be appropriate in that position. If it is, two partial segmentation candidates may be generated, with and without the boundary. At some point, the number of candidates exceeds some predefined upper limit $n$, the beam width, and the $n$ best candidates are chosen to continue the process. When the whole input is processed by this algorithm, $m$ candidates from the currently $n$ best are chosen as the final result. In Figure 1 we show a simplification of the described procedure.

All decisions made by the algorithm are based on the information retrieved from the language model, which estimates the likelihood of sequences of tokens. More precisely, given an input token and a history of $\rho$ previous tokens, the language model approximates the probability distribution over all the possible token values for the next token in the sequence.

In the following sections we describe each of these components in more detail.

### Language Model

We implemented the language model that provides all the necessary information for the search algorithm as a recurrent neural network (Mikolov & Zweig, 2012) and an n-gram model (Heafield, Pouzyrevsky, Clark, & Koehn, 2013).

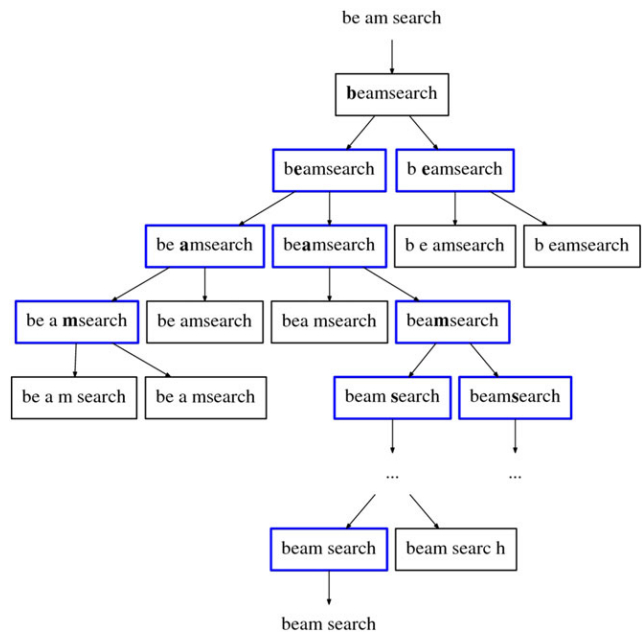

FIG. 1. Simplified illustration of the algorithm execution, with $n = 2$ and $m = 1$. [Color figure can be viewed at wileyonlinelibrary.com]

In essence, this information is the probability of the occurrence of some input token $x_t$, usually in the logarithmic scale, given the previous $\rho$ tokens in the input:

$$logP(x_t) = logP(x_t|x_{t-1},\ldots,x_{t-\rho}) \quad (1)$$

For the neural model, $\rho$ is defined at training time as a hyperparameter of the neural network used for the Truncated Back-Propagation Through Time (TBPTT; Principe, Principe, & Kuo, 1993). In the case of the n-gram model, $\rho = n - 1$, where $n$ is the order of the model, that is, the number of tokens in each n-gram.

With this information we can also obtain the estimated likelihood of an input token sequence as the mean of the probabilities of its constituent tokens:

$$score(x_t, x_{t-1},\ldots,x_0) = \frac{1}{t}\sum_{n=0}^{t} logP(x_n) \quad (2)$$

The choice of using a recurrent neural network is based on its focus on dealing with sequential data, such as text, as well as its wide use in several NLP tasks, such as machine translation (Johnson et al., 2016), dependency parsing (Vilares & Gómez-Rodríguez, 2017), question answering (Iyyer, Boyd-Graber, Claudino, Socher, & Daumé III, 2014), or language modeling (Sundermeyer, Schlüter, & Ney, 2012). Recurrent neural networks differ from traditional feedfoward networks in that they allow feedback loops in their architectures, thus being able to use the output information corresponding to the input $t$ when processing input $t + 1$. The existence of different types of recurrent networks comes from the different designs of their recurrent units. In our case, we used LSTM units (Hochreiter & Schmidhuber, 1997) for the construction of our neural networks, as they have proven very effective for language modeling (Sundermeyer et al., 2012). These LSTM units contain a memory cell, which stores information from past computations, and three *gates* that control the information stored in the memory as well as the output of the whole unit.

Using a byte-level approach we can reuse the same network design for multiple languages, as the character set is not a parameter in the design process. This can also be an advantage for languages with large character sets, as fixing a smaller output size for the softmax operation in the last layer of the network avoids the bottleneck issues caused by this operation on large vocabularies. Furthermore, to reduce even more the complexity of the problem, we do not consider those byte values corresponding to nonprintable characters except for the null byte 0, which can be used as padding in the input sequence. Assuming a Unicode encoding such as the popular UTF-8, these are the values in the range [1, 31]. The resulting neural networks receive as input one byte at a time from a given sequence and output the (logarithmic) probabilities for each of the possible next bytes in the sequence. The general architecture of these networks is depicted in Figure 2.

An n-gram model is a structure that stores historic data about the n-grams, sequences of *n* tokens, seen in a training corpus. For constructing these models, we first have to set the order of the model, *n*, and usually the smoothing function, although in this case the Kneser-Ney smoothing (Heafield et al., 2013; Kneser & Ney, 1995) seems to be the best option overall (Chen & Goodman, 1996).

We avoid the data sparsity problem mentioned in the Introduction by using language models that work at the byte and character level instead of the word level, and also by using neural networks. These latter models transform the sparse discrete input data, usually *one-hot* vectors, into continuous representations that encode meaningful information about the relations between the inputs and outputs of the network (Kim et al., 2015). Under this assumption, three words such as hiiii, hii, and hi would end up having *similar* continuous representations, as they are morphologically similar and would appear in similar contexts.

However, as powerful constructions as recurrent neural networks are, they tend to overfit the training data (Zaremba, Sutskever, & Vinyals, 2014). To overcome this issue, several measures may be used, of which we have chosen Batch Normalization (Ioffe & Szegedy, 2015) and keeping the network as small as possible while retaining a good precision in the task at hand.

*Beam Search Algorithm*

Now we describe the beam search algorithm using a functional approach. For all the following functions, we define the threshold parameter $t$, beam width $b$, number of final results $m$, word boundary element $wb$, and scoring function *score* as global constants in order to avoid long function signatures and improve readability. In addition, the $+$ and $\oplus$ symbols are used as the operators for string and list concatenation, respectively, $s_i$ denotes the character at position $i$ from the string $s$, $s_{i,j}$ denotes the substring of $s$ going from index $i$ to $j$, and $l_{i,j}$ the sublist of $l$ going from index $i$ to $j$.

The first function we define is $segment^*(part, txt)$, which recursively processes one token at a time from the input sequence. It takes two arguments: A list of partial results for the already-processed text and the remaining text to segment with no word boundaries and length $l$. The path this function takes depends on the emptiness of its first and last arguments. If the first one is $\emptyset$, we have the base case and the recursion stops returning the current partial results list. If the second argument is $\emptyset$, the function bootstraps a partial results list and calls itself appropriately to begin the recursive process. This is the way that the function should be called the first time. The remaining case is the main recursive case, where the function calls *beam*.

$$segment^*(part, txt) := \begin{cases} part & txt = \emptyset \\ segment^*((txt_1), txt_{2,l}) & part = \emptyset \\ segment^*(beam(part, txt_1), txt_{2,l}) & otherwise \end{cases}$$
(3)

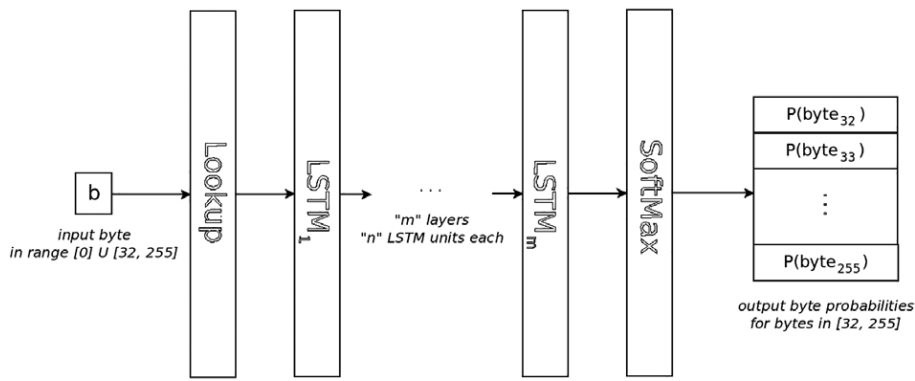

FIG. 2. Illustration of the architecture of our neural networks. The *Lookup* layer transforms a number (byte) into a tensor suitable for the first LSTM layer. The *LSTM* layers apply a nonlinear transformation to their inputs. The *SoftMax* layer computes the output probabilities using the output from the last LSTM layer.

*beam(c, part)* takes as a first argument the current token in the sequence and obtains a new list of *n* best partial results including that token in the segmentation. For this, it calls the *xpd(part, c)* (expand) and $top_n(part, b)$ functions.

$$beam(part, c) := top_n(xpd(part, c), b) \quad (4)$$

$$xpd(part, c) := \begin{cases} \emptyset & part = \emptyset \\ (part_1 + c) \oplus (bnd(part_1, c)) \oplus xpd(part_{2,|part|}, c) & otherwise \end{cases} \quad (5)$$

The first one obtains new candidates by recursively traversing the partial results list and generating at least one more candidate for each of the existing ones by appending the next token *c* to them. It may also generate a second candidate if the call to the function *bnd(x, c)* (boundary) does not return $\emptyset$.

$$bnd(x, c) := \begin{cases} x + wb + c & score(x + wb + c) > t \\ \emptyset & otherwise \end{cases} \quad (5)$$

This function checks, using the scoring function *score*, if a candidate *x* with a word boundary in the last position followed by *c* is likely or not. This is, if its associated score (likelihood as returned by the scoring function) is greater than the threshold parameter *t*. If it is, it will return this new candidate.

The $top_n$ function selects the *n* best partial results.

$$top_n(part, n) := sort((x, y) \mapsto score(x) > score(y), part)_{1,n} \quad (6)$$

The final step would be to create a wrapper function that acts as an entry point to the system through the correct call to *segment*. This function will also call *r* to remove word boundary characters from the input text and select the top *m* partial results to serve as final results.

$$segment(txt, t, b, wb, score, m) = top_n(segment^*(\emptyset, r(txt)), m) \quad (7)$$

$$r(str) := filter((x) \mapsto x \neq wb, str) \quad (8)$$

### Experiments

In this section we describe the implementation details and followed procedure for validating our approach.

*System Implementation*

We implemented two versions of the system just described, one in Lua and the other in Python, due to the availability of the tools that we use to implement the language models. Torch,[4] a scientific framework with support for neural networks, is available for Lua, and kenlm,[5] a toolkit for n-gram language modeling, has bindings for easy usage in Python.

Torch only includes by default the tools to build feed-forward neural networks, so in order to use it for recurrent neural networks we imported the package rnn (Léonard, Waghmare, Wang, & Kim, 2015). We also used the Adam optimization algorithm (Kingma & Ba, 2014) from the optim[6] Lua package.

On the other hand, the kenlm toolkit is straightforward to use. The generation of the n-gram models was performed from the command line, while their integration with the search algorithm took place inside a Python script.

All the implemented code is available at http://www.grupocole.org/software/VCS/segmnt/.

For the training and evaluation of the neural models, we tried to take full advantage of the parallelization features of Torch. Thus, all computations are performed in batches by a GPU, in our case a GTX Titan X (2015).

Another implementation detail not previously specified is the extra numeric parameter *win*. It defines the number of previous tokens from the current position in the input to use for the score computation. This is, instead of computing $score(x_t, x_{t-1}, ..., x_0)$ as shown in Equation 2, we compute $score(x_t, x_{t-1}, ..., x_{t-win})$. As this scoring operation is costly for the neural language model, this parameter allows us to seek a compromise between execution time and accurate scoring. It is worth noting that the value assigned to *win* does not have to be necessarily the same as the one used for $\rho$ (see Language model).

*Corpora*

The data used for training the models, both for our system and WordSegment, was obtained from several sources. For English, German, Turkish, and Finnish, we used the monolingual training data sets corresponding to the 2016 news from the WMT17 shared task, available at http://www.statmt.org/wmt17/. The English corpus was also augmented with tweets from the training data set at http://cs.stanford.edu/people/alecmgo/trainingandtestdata.zip.

For each one of our data sets, we shuffled the lines[7] and selected the first 10 million lines (at most) for training and the last 300 (600 in the case of English) for validation. We then removed special tokens such as microblog mentions, hashtags, and URLs, as they constitute counter-examples of the tokens we want to obtain in our results. For the Finnish and Turkish data sets, we also removed the SGML tags and some resulting blank lines.

---

[4] http://torch.ch/
[5] https://kheafield.com/code/kenlm/
[6] https://github.com/torch/optim
[7] A line is defined as a sequence of characters delimited by newline characters.

In the case of Spanish, we used the same training corpus as in our previous work (Doval et al., 2016), which is based on the Wikipedia dump from 2015/02/28 preprocessed using the wikiextractor[8] and with all the Wikipedia markup expressions removed. From the result, we selected the first 4 million lines.

As test data, we used the monolingual testing data sets corresponding to the 2013 news from the WMT17 shared task for English and German, and the ones corresponding to 2016 for Turkish and Finnish, as there is no test data from 2013 available for these languages. The preprocessing performed was the same as described above for the training corpus. The difference here is that we kept the English tweets test corpus, obtained from the same source as the training corpus, separated from the news test corpus. It is also important to note that, unfortunately, we did not have enough resources available at the time to normalize some aspects of the tweets used for testing. These cause a correct segmentation to be labeled as incorrect, such as with the output "no way" for the reference "noway." This has a greater impact on our approach and WordSegment than on the Word Breaker, as it considers a wider range of input characters, which may be incorrectly positioned in the reference. As an example, the "-" character may be particularly difficult to test properly as it tends to appear arbitrarily surrounded by whitespaces in informal contexts.

To alleviate this and provide a fair comparison, we add a particular test case for these systems where we only consider the correct positioning of word boundaries around alphanumeric tokens. We also give the precision score for the corresponding *strict* test case where we consider the positioning of all word boundaries.

For Spanish we used two test data sets. One of them is the same as in our previous work (Doval et al., 2016), based on the same Wikipedia article dump used for training where we randomly select 1,000 short lines from the last 25% of the lines in the corpus. It is used here to facilitate the comparisons between our current approach and our previous one. The second one is obtained in almost the same way as the former, but in this case the lines are kept noticeably longer, as the random selection considers lines regardless of their length. Given the exact match scoring scheme that we will use, this should pose a greater challenge for all the systems in the benchmark.

All data set preprocessing scripts are available at http://www.grupocole.org/software/VCS/segmnt/.

---

[8] https://github.com/attardi/wikiextractor

## Results

In our tests, we focus on precision as the performance metric. Precision is defined here as the number of correctly segmented input instances over the total number of inputs given to the system. A correct segmentation means that every word boundary in the output is correctly placed, otherwise it is deemed incorrect; that is, our precision metric is an exact match score over whole lines.

We began our experiments by training some neural models and checking their performance in the development set throughout this process. We tried smaller networks first and progressively increased their parameter count until we reached bigger models with reasonable size and performance, both in precision and time metrics. Due to the high costs of training neural networks, which usually took days to complete, we were far from exhaustive in our exploration of the hyperparameter space. For this same reason, these initial experiments were only conducted with the English and Twitter corpora. The best models for English and Twitter would then be used for the remaining languages.

In Figure 3 we show validation error curves for a few relevant models and in Table 2 the precision numbers obtained by those in the segmentation task. The results obtained for the tweets *strict* case are around 20 points below those shown in the table (see Corpora). As we expected, lower validation error numbers can be obtained with bigger networks, which generally translates into higher precision figures.

For standard text, relatively good precision can be obtained even by the smallest network. If we want to see the real benefits of more complex networks, we have to look at the precision numbers for a more difficult setup such as the tweets data set. These networks, containing a higher number of parameters, are better suited for handling the greater sparseness in microtext data.

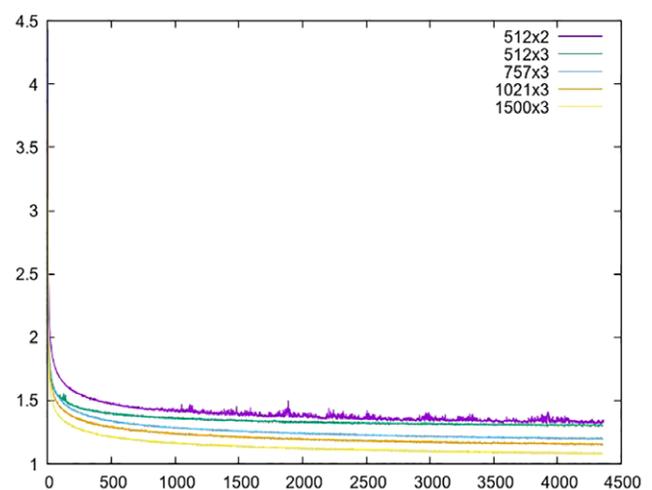

FIG. 3. Validation error curves for some neural models on the English and Twitter training corpus. The architecture is specified as $n \times m$, where $m$ is the number of hidden layers and $n$ the number of neurons per layer. [Color figure can be viewed at wileyonlinelibrary.com]

TABLE 2. Precision results on the English and Twitter development data sets by language and neural model architecture specified as $n \times m$, where $m$ is the number of hidden layers and $n$ the number of neurons per layer.

|  | English | Tweets |
|---|---|---|
| $512 \times 2$ | $0.820_{2510}$ | $0.756_{1299}$ |
| $512 \times 3$ | $0.823_{3127}$ | $0.783_{1472}$ |
| $757 \times 3$ | $0.820_{3766}$ | $0.790_{1800}$ |
| $1021 \times 3$ | $0.823_{4482}$ | $0.813_{2571}$ |
| $1500 \times 3$ | **$0.866_{8235}$** | **$0.826_{4000}$** |

Elapsed times in seconds are also shown in subscript.

However, it seems that our networks cannot grow indefinitely in width (number of neurons per layer) with respect to depth (number of layers), as this may cause serious unstability issues during the training process (see Figure 4).

Consequently, in order to insert more neurons per layer, it would also be necessary to insert more layers into the network at some point. Moreover, given the long training process required by these bigger models and the precision numbers obtained, which are shown below, we decided to stop at networks of three layers and 1,500 neurons per layer.

Next, we built our n-gram models, a process that took minutes even for the largest of these models.

The only mandatory parameter we had to adjust here was the order of the model, $n$, for which we considered the values 4, 6, 8, 10, and 12. The precision numbers for each of these models on the development data sets can be seen in Table 3. In this case, the results obtained in the *strict* test are around 30 points below those shown in the table.

As with the number of parameters of neural models, our n-gram models also benefit from higher values of $n$. After the large performance gain when moving from order 4 to order 6, the growth slows down until it stops or reverses for most languages when going from order 10 to order 12. On the other hand, these models are usually notably affected by the sparsity of the training data, and using higher $n$ values yields exponentially larger models. To avoid this, we can prune those n-grams with frequency counts lower than a specified threshold value at the expense of possibly lowering the performance of the model. In our tests, we used a pruning value of 5 for the 8-, 10-, and 12-grams without any noticeable performance drop.

For the reasons explained above, we decided to stop constructing bigger models at $n = 12$. This is similar to the reasoning applied in the case of neural models, adding the relatively small performance gain when going from $n = 10$ to $n = 12$.

In order to account for the difference in execution time for different languages, we show in Table 4 the average counts of words, characters, and bytes per line (instance) in the development data sets.

Before obtaining the precision figures presented, we also tuned the search algorithm parameters for each type of language model using the development data sets. For the neural model, these were $t = 8$, $b = 10$, $win = 64$, whereas the n-gram model required $t = 10$, $b = 500$, $win = \infty$ to guarantee good performance across data sets.

These numbers imply that the neural model is able to make better decisions at each step of the segmentation algorithm and thus requires carrying fewer partial candidates in order to finally decide on a good one. The opposite seems to be true for the n-gram model, which needs a wide range of possibilities available at each step and thus a higher $b$ value, in proportion with the length of the text input.

Then we compared our system, WordSegment, and the Microsoft Word Breaker (as of April 2017) against each other using the test data sets. In order to interact with the latter system, we used a slightly modified version of the demo code available at its website, transcribing or removing non-ASCII characters and adapting the formatting of the output to make it compatible with the rest of our evaluation scripts. The best precision numbers on the development data sets were obtained with $n = 3$ and the Bing body corpus. It is worth noting that, as this system works only with alphanumeric characters, the test gold standard was filtered accordingly. This implies that the failure surface for the Word Breaker is much smaller compared to that of our systems and WordSegment, as the former does not consider possibly troublesome characters such as "'/" or "-".

The results are shown in Table 5, where we can see that both the n-gram and neural models were able to obtain higher precision numbers than both WordSegment and Word Breaker on the test data sets. The sole exception is the tie between our 12-gram model and WordSegment in the smaller Spanish data set, which is resolved in our favor in the longer Spanish data set. For further detail, we have published more complete outputs at http://www.grupocole.org/software/VCS/segmnt/.

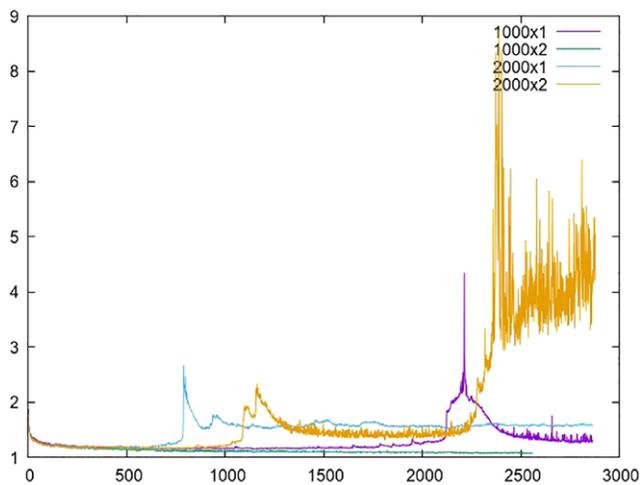

FIG. 4. Validation error curves for models that are *too* wide. The architecture is specified as $n \times m$, where $m$ is the number of hidden layers and $n$ the number of neurons per layer. [Color figure can be viewed at wileyonlinelibrary.com]

TABLE 3. Precision results on development data sets by language and n-gram model order.

|        | EN          | DE          | TR          | FI          | Tw          | ES          |
|--------|-------------|-------------|-------------|-------------|-------------|-------------|
| $n=4$  | $0.376_{57}$ | $0.340_{47}$ | $0.513_{49}$ | $0.336_{30}$ | $0.486_{18}$ | $0.370_{279}$ |
| $n=6$  | $0.866_{68}$ | $0.773_{55}$ | $\mathbf{0.940}_{54}$ | $0.820_{33}$ | $0.773_{25}$ | $0.726_{344}$ |
| $n=8$  | $0.890_{70}$ | $0.846_{59}$ | $0.926_{56}$ | $0.883_{35}$ | $0.776_{25}$ | $0.780_{386}$ |
| $n=10$ | $\mathbf{0.906}_{87}$ | $\mathbf{0.893}_{72}$ | $0.930_{61}$ | $0.916_{41}$ | $\mathbf{0.793}_{32}$ | $\mathbf{0.793}_{425}$ |
| $n=12$ | $0.893_{105}$ | $\mathbf{0.893}_{92}$ | $0.930_{65}$ | $\mathbf{0.926}_{49}$ | $0.790_{41}$ | $\mathbf{0.793}_{451}$ |

Elapsed times in seconds are also shown in subscript.

TABLE 4. Average counts of words, characters, and bytes per instance in the development data sets.

|            | EN     | DE     | TR     | FI    | Tw    | ES     |
|------------|--------|--------|--------|-------|-------|--------|
| words      | 20.00  | 15.24  | 13.58  | 10.04 | 13.07 | 42.14  |
| chars      | 120.24 | 106.48 | 103.40 | 90.33 | 74.06 | 260.96 |
| bytes      | 120.24 | 107.95 | 112.91 | 93.52 | 74.07 | 266.28 |
| chars/word | 6.01   | 6.98   | 7.61   | 8.99  | 5.66  | 6.19   |
| bytes/char | 1.00   | 1.01   | 1.09   | 1.03  | 1.00  | 1.02   |

WordSegment suffers from data sparsity problems, as we can infer from the heavy performance drop in the Turkish, Finnish, and Twitter test corpora. For Word Breaker, the performance gap was smaller in the case of standard English and Spanish. We believe that this is due to the language model of Word Breaker having seen many more tokens from these languages than from German, Turkish, or Finnish, where it performed noticeably worse than its competitors.

More interestingly, we also come up well above in the English tweets data set, a particularly challenging domain. This time, the reason might be the acute data sparsity present in this kind of texts, whose word vocabulary includes a wide range of texting-induced variations of standard words, which are handled well by the character-based approach.

Regarding the neural models, we can see that our currently biggest network does not perform much better than the second biggest across the test bench. In some cases, such as with Turkish and, most notably, Spanish, it is actually worse.

We can also observe that the performance of the n-gram models was close to the neural models, most notably for the Finnish language, and even surpassed them by a noticeable margin in the case of Spanish. Given the great attention and good results obtained by neural models in the literature, we expected the opposite to be true. To add more merit to the n-gram models, we should also mention their (quite) faster operation, both in training and evaluation time, compared with the neural models.

The main reasons why we did not try bigger or more sophisticated neural models are the good results we already achieved and the long training processes and slow operation times we obtain with our current biggest models, even on such a powerful GPU as the Titan X (2015).

We also observed that it may be possible to apply this approach in a cross-lingual environment, even though our training and testing corpora are monolingual. The short English phrase formed by common words "You are not welcome" in the Finnish test corpus is correctly segmented, as well as English named entities in the German and Turkish corpora such as "The Particle Adventure" or "The Voice." It is then reasonable to assume that, given a suitable cross-lingual training corpus, it should be possible to address a truly cross-lingual scenario where sentences mix words from different languages.

Finally, our new approach clearly improves on the previous one proposed (Doval et al., 2016), which obtained 0.82 in the shorter Spanish test data set, taking more than twice the time taken by our currently biggest neural model.

## Conclusion

In this work we presented a new approach to tackle the word segmentation problem consisting of two components: A beam search algorithm, which generates and chooses over possible segmentation candidates incrementally while scanning the input one token at a time, and a language model working at the byte or character level, which enables the algorithm to rank those candidates. We considered recurrent neural networks and n-gram models to implement the language model. This work is a continuation of Doval et al. (2016).

TABLE 5. Precision results on the test data sets by language and approach.

|                      | EN    | DE    | TR    | FI    | Tw    | ES $S$ | ES $L$ |
|----------------------|-------|-------|-------|-------|-------|-------|-------|
| 10–gram              | 0.914 | 0.822 | 0.776 | 0.775 | 0.757 | 0.924 | **0.803** |
| 12–gram              | 0.913 | 0.824 | 0.775 | **0.784** | 0.765 | **0.927** | 0.771 |
| $1021 \times 3$–neural | 0.904 | **0.885** | **0.823** | 0.753 | **0.781** | 0.894 | 0.759 |
| $1500 \times 3$–neural | **0.922** | **0.885** | 0.811 | 0.773 | **0.781** | 0.867 | 0.765 |
| WordSegment          | 0.866 | 0.720 | 0.551 | 0.577 | 0.542 | **0.927** | 0.616 |
| Word Breaker         | 0.887 | 0.385 | 0.154 | 0.089 | 0.626 | 0.880 | 0.665 |

Our aim was to build a word segmentation system that can be used in the context of microtexts, a domain where data sparsity can be a problem to traditional approaches based on word n-grams, such as the popular Microsoft Word Breaker or the WordSegment Python module. We solve this issue by using byte- and character-level language models, and also by taking advantage of the ability of neural networks to transform their discrete sparse inputs into continuous representations that encode similarities between inputs.

In our experiments, we explored possible configurations for our systems by adjusting the search algorithm parameters and the language model hyperparameters. The languages we considered for training and testing were English, German, Turkish, Finnish, Spanish, and also English tweets. Then we compared the performance of the different configurations of our system, WordSegment, and the Microsoft Word Breaker. The best neural models obtain the best precision figures overall on the test data sets.

Surprisingly, the performance of the simpler n-gram models was close to their neural counterparts while being noticeably faster. Compared to WordSegment and the Word Breaker, our approach obtained better results overall.

We expect that the advancements that are rapidly taking place in neural network frameworks and parallel architectures will allow us to further improve the execution time and performance of the neural models with little or no changes to their current core design.

Aside from this, as future lines of work we plan on integrating this system into a microtext normalization pipeline as a preprocessing step. We may also see how this solution fares in the more studied context of Asian languages, mainly Chinese.

## Acknowledgments

This research received funding from the European Research Council (ERC) under the European Union's Horizon 2020 research and innovation program (grant agreement no. 714150-FASTPARSE). It was partially funded by the Spanish Ministry of Economy and Competitiveness (MINECO) through projects FFI2014-51978-C2-1-R and FFI2014-51978-C2-2-R, and by the Autonomous Government of Galicia through both the Galician Network for Lexicography-RELEX (ED431D R2016/046) and Grant ED431B-2017/01. Moreover, Yerai Doval is funded by the Spanish State Secretariat for Research, Development and Innovation (which belongs to MINECO) and by the European Social Fund (ESF) under an FPI fellowship (BES-2015-073768) associated with project FFI2014-51978-C2-1-R. We gratefully acknowledge NVIDIA Corporation for the donation of a GTX Titan X GPU used for this research.## References

Adda-Decker, M., Adda, G., & Lamel, L. (2000) Investigating text normalization and pronunciation variants for German broadcast transcription. In Proceedings of the Sixth International Conference on Spoken Language Processing, ICSLP 2000 INTERSPEECH 2000 (pp. 266–269), Beijing, China. Baixas, France: ISCA.

Alfonseca, E., Bilac, S., & Pharies, S. (2008). Decompounding query keywords from compounding languages. In Proceedings of the 46th Annual Meeting of the ACL: Short Papers (pp. 253–256). Stroudsburg, PA: ACL. Retrieved from http://dl.acm.org/citation.cfm?id=1557690.1557763

Cai, D. & Zhao, H. (2016). Neural word segmentation learning for Chinese. In Proceedings of the 54th Annual Meeting of the ACL 2016 (Vol. 1), Berlin, Germany. Stroudsburg, PA: ACL. Retrieved from http://aclweb.org/anthology/P/P16/P16-1040.pdf.

Chen, S.F. & Goodman,J. (1996). An empirical study of smoothing techniques for language modeling. In Proceedings of the 34th Annual Meeting on ACL 1996, Santa Cruz, CA (pp. 310–318). Stroudsburg, PA: ACL.

Chen, X., Qiu, X., Zhu, C., & Huang, X. (2015). Gated recursive neural network for Chinese word segmentation. In Proceedings of the 53rd Annual Meeting of the ACL and the Seventh International Joint Conference on Natural Language Processing of the Asian Federation of Natural Language Processing 2015, Beijing, China (Vol. 1, pp. 1744–1753). Stroudsburg, PA: ACL. Retrieved from http://aclweb.org/anthology/P/P15/P15-1168.pdf

Chen, X., Qiu, X., Zhu, C., Liu, P., & Huang, X. (2015). Long short-term memory neural networks for Chinese word segmentation. In Proceedings of the Conference on Empirical Methods in Natural Language Processing (EMNLP 2015), Lisbon, Portugal (pp. 1197–1206). Stroudsburg, PA: ACL.

Chi, C.-H., Ding, C., & Lim, A. (1999). Word segmentation and recognition for web document framework. In Proceedings of the Eighth International Conference on Information and Knowledge Management (CIKM '99), Kansas City, MO (pp. 458–465). New York: ACM. Retrieved from http://doi.acm.org/10.1145/319950.320051

Doval, Y., Gómez-Rodríguez, C., & Vilares, J. (2016). Segmentación de palabras en español mediante modelos del lenguaje basados en redes neuronales. Procesamiento del Lenguaje Natural, 57, 75–82.

Doval, Y., Vilares, J., & Gómez-Rodríguez, C. (2015). LYSGROUP: Adapting a Spanish microtext normalization system to English. In ACL 2015 Workshop on Noisy User-generated Text, Proceedings of the Workshop (pp. 99–105), Beijing, China. Red Hook, NY: Curran Associates.

Heafield, K., Pouzyrevsky, I., Clark, J.H., & Koehn, P. (2013). Scalable modified Kneser-Ney language model estimation. In Proceedings of the 51st Annual Meeting of the ACL 2013, Sofia, Bulgaria (pp. 690–696). Stroudsburg, PA: ACL. Retrieved from http://kheafield.com/professional/edinburgh/estimate_paper.pdf

Hochreiter, S., & Schmidhuber, J. (1997). Long short-term memory. Neural Computation, 9(8), 1735–1780.

Ioffe, S. & Szegedy, C. (2015). Batch normalization: Accelerating deep network training by reducing internal covariate shift. In Proceedings of the 32nd International Conference On Machine Learning (ICML-15; pp. 448–456). Lille, France.

Iyyer, M., Boyd-Graber, J.L., Claudino, L.M.B., Socher, R., & Daumé III, H. (2014). A neural network for factoid question answering over paragraphs. In Proceedings of the Conference on Empirical Methods in Natural Language Processing (EMNLP 2014; pp. 633–644). Stroudsburg, PA: ACL.

Johnson, M., Schuster, M., Le, Q.V., Krikun, M., Wu, Y., Chen, Z., ..., Dean, J. (2016). Google's multilingual neural machine translation system:10    JOURNAL OF THE ASSOCIATION FOR INFORMATION SCIENCE AND TECHNOLOGY—Month 2018
       DOI: 10.1002/asi